\documentclass[sigconf,10pt]{acmart}
\renewcommand\footnotetextcopyrightpermission[1]{}
\makeatletter
\DeclareRobustCommand\onedot{\futurelet\@let@token\@onedot}
\def\@onedot{\ifx\@let@token.\else.\null\fi\xspace}

\makeatother

\newcommand{\ignore}[1]{}   

\newcommand{\lidar}[1]{LIDAR}
\usepackage{enumitem}
\usepackage[T1]{fontenc}
\usepackage{algorithm}
\usepackage[noend]{algpseudocode}

\usepackage{tikz}
\usetikzlibrary{shapes,arrows.meta,positioning,fit,calc}

\usepackage{subcaption} 

\usepackage{xcolor}
\newcommand{\best}[1]{\textbf{#1}}
\newcommand{\down}{\ensuremath{\downarrow}}
\newcommand{\up}{\ensuremath{\uparrow}}

\author{Gaurav Bagwe}
\affiliation{%
  \institution{Clemson University}
  \city{Clemson}
  \state{South Carolina}
  \country{USA}
}
\email{gbagwe@clemson.edu}

\author{Xiaoyong (Brian) Yuan}
\affiliation{%
  \institution{Clemson University}
  \city{Clemson}
  \state{South Carolina}
  \country{USA}
}
\email{xiaoyon@clemson.edu}

\author{Yongji Wu}
\affiliation{%
  \institution{Clemson University}
  \city{Clemson}
  \state{South Carolina}
  \country{USA}
}
\email{yongjiw@clemson.edu}

\author{Lan Zhang}
\affiliation{%
  \institution{Clemson University}
  \city{Clemson}
  \state{South Carolina}
  \country{USA}
}
\email{lan7@clemson.edu}
\usepackage{array}
\usepackage{multirow}
\usepackage{makecell}

\usepackage{booktabs}

\newcolumntype{H}{>{\hspace{0pt}\ignorespaces}p{0pt}<{\unskip\hspace{0pt}}}
\begin{document}
\title{
DiffRadar: Differentiable Physics-Aware Radar SLAM with Gaussian Fields
}

\begin{abstract}
Radar sensing is increasingly used in mobile systems because it operates reliably under poor lighting, adverse weather, and privacy-sensitive settings where cameras and LiDAR often fail. However, most existing radar SLAM systems estimate motion through scan matching on discretized radar heatmaps, which breaks geometric continuity and fails to capture key radar sensing properties, often leading to unstable pose estimation and degraded mapping in degenerate or dynamically changing environments. We present \textbf{DiffRadar}, a real-time radar SLAM system that models radar observations as a differentiable, physics-aware Gaussian field rather than discrete scans. 
DiffRadar represents the scene as anisotropic Gaussian primitives and renders radar measurements in range–azimuth and Doppler–azimuth spaces through a differentiable radar forward model, enabling joint optimization of robot pose and scene structure directly from radar measurements. 
We implement DiffRadar on commodity FMCW radar hardware and evaluate it on both the public Radarize benchmark and a controlled stress-test suite that targets common radar SLAM failure modes, including corridor degeneracy, motion regime transitions, dynamic clutter, and long-horizon loop closures. 
DiffRadar achieves substantial reductions in trajectory error on the benchmark, with especially large gains under feature-poor corridor motion, while more than doubling map consistency and maintaining real-time performance at 70 FPS. These results show that modeling radar observations directly in the signal domain enables substantially more robust and consistent radar-only SLAM for mobile platforms.

\end{abstract}



\maketitle

\section{Introduction}
Simultaneous Localization and Mapping (SLAM) is a fundamental capability for mobile systems operating in unfamiliar environments~\cite{placed2023survey,ress2024slam,taketomi2017visual,eliazar2003dp}. Autonomous robots, vehicles, and edge devices rely on SLAM to continuously estimate their positions while constructing a map of their surroundings~\cite{montemerlo2002fastslam,eliazar2003dp}. Most modern SLAM systems rely on optical sensors such as cameras or LiDAR, which can achieve centimeter-level accuracy in structured scenes~\cite{xu2022swarmmap,matsuki2024gaussian}. However, these sensors degrade significantly under adverse sensing conditions, including poor lighting, fog, glare, or privacy-sensitive environments where optical sensing may be undesirable~\cite{brune2024survey,pinchon2018all}. 
Enabling reliable SLAM under such conditions remains a key challenge for mobile and wireless systems.

Millimeter-wave (mmWave) radar provides a promising sensing modality for robust localization in these scenarios. Radar sensing is largely invariant to lighting conditions and can penetrate common obscurants such as smoke, dust, and fog~\cite{lu2020see,lu2020milliego,sie2024radarize}. Modern mmWave radars are also compact, inexpensive, and increasingly integrated into commodity mobile platforms, making radar an attractive option for RF-based spatial perception when optical sensing is unreliable.

However, radar perception differs fundamentally from optical sensing. Radar observations are sparse, noisy, and anisotropic, providing fine resolution in range but limited resolution in azimuth. 
Existing radar SLAM systems~\cite{sie2024radarize,lu2020milliego,prabhakara2023high,li2025super4dr}
typically process radar observations as discretized range–azimuth (RA) heatmaps and estimate motion through modular pipelines such as scan matching or Doppler-based odometry.
While effective in some environments, this abstraction limits the ability to exploit the underlying physics of radar sensing.
Discretizing radar observations breaks geometric continuity and often treats radar intensity as isotropic, ignoring directional backscatter and Doppler structure.
Consequently, motion estimation and mapping are handled in separate modules, reducing robustness when individual sensing cues become unreliable.

\begin{figure}[t]
  \centering
  \includegraphics[width=0.49\textwidth]{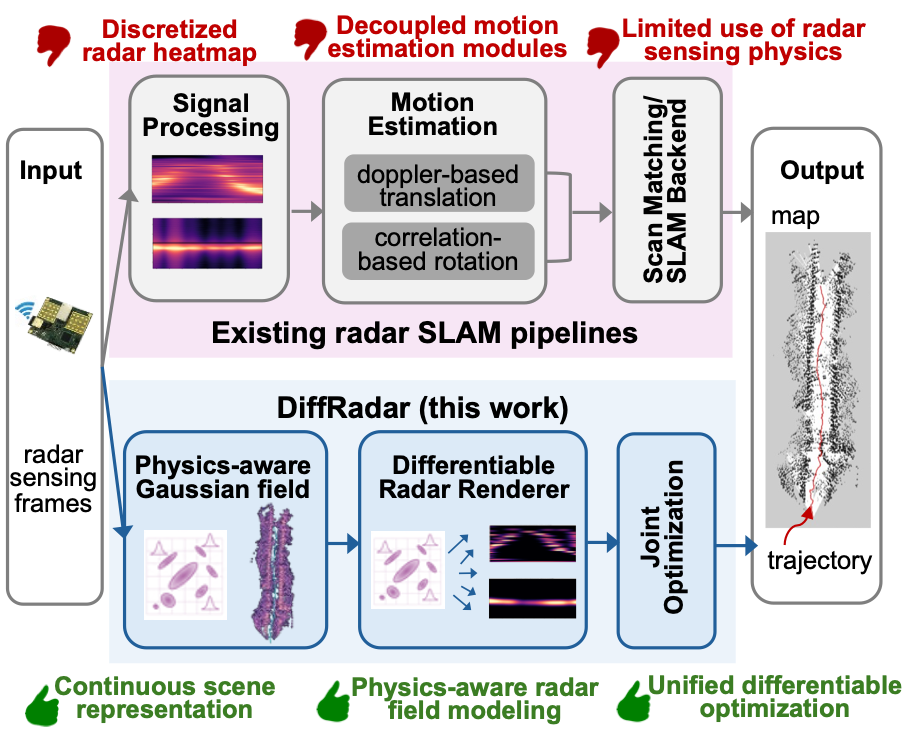}
\caption{From modular radar pipelines to differentiable radar SLAM. 
Existing radar SLAM systems process discretized radar heatmaps and estimate motion through separate modules. 
DiffRadar instead models radar observations as a continuous physics-aware Gaussian field and performs joint pose-map optimization through differentiable rendering.} \label{fig:radar_limiations}
\Description{Block diagram comparing conventional radar SLAM with DiffRadar. The top pipeline shows radar frames converted to discretized heatmaps, followed by separate motion-estimation modules and scan matching to produce a map. The bottom pipeline shows DiffRadar representing the scene as a continuous physics-aware Gaussian field, rendering radar measurements differentiably, and jointly optimizing map and trajectory in one loop.}

\end{figure}

At the core of these limitations lies a fundamental abstraction mismatch:
radar observations are typically treated as \emph{images} to be processed rather than \emph{physical signals governed by radar sensing physics}.
This observation suggests formulating radar SLAM directly in the radar signal domain rather than through discretized image representations.
In particular, radar-specific properties such as anisotropic measurement uncertainty, directional scattering, and Doppler dynamics are difficult to incorporate within image-based pipelines.
Figure~\ref{fig:radar_limiations} illustrates this abstraction mismatch.

To address this abstraction mismatch, we propose \textbf{DiffRadar}, a real-time radar SLAM system that models radar observations through a differentiable, physics-aware Gaussian field. 
Unlike existing radar SLAM pipelines that operate on discretized radar heatmaps, DiffRadar represents the environment as a continuous field of anisotropic Gaussian primitives and jointly optimizes robot pose and scene structure through a radar-domain forward model. Each primitive encodes scene geometry and radar sensing characteristics, enabling radar measurements to be rendered in RA and DA spaces.

DiffRadar integrates three key components.
\textit{(1) Physics-aware Gaussian representation} models radar scatterers as anisotropic Gaussian primitives whose spatial structure reflects radar sensing characteristics such as range resolution and angular uncertainty.
\textit{(2) Differentiable radar renderer} projects the Gaussian field into RA and DA measurement spaces, enabling gradients to propagate from measurement residuals to scene and pose parameters.
\textit{(3) Unified pose–map optimization} jointly refines the robot trajectory and scene representation by minimizing discrepancies between rendered and measured radar observations.

By grounding radar SLAM in sensing physics, DiffRadar replaces modular heatmap-based processing with a unified differentiable optimization framework that yields stable motion estimation and consistent mapping even in regimes where traditional radar SLAM pipelines are prone to drift or degeneracy. 
Our contributions are summarized as follows:
\begin{itemize}
\item \textbf{Continuous physics-aware radar representation.}
We introduce a Gaussian field representation for radar SLAM that models radar observations as continuous primitives rather than discretized heatmaps, enabling smooth geometric modeling and gradient-based optimization over scene structure and motion.

\item \textbf{Differentiable radar renderer.}
We design a radar-domain differentiable renderer that projects the Gaussian field into RA and DA measurement spaces while preserving radar sensing physics, allowing measurement residuals to propagate directly to scene and pose parameters.

\item \textbf{Unified pose-map optimization.}
We develop a joint optimization framework that refines robot trajectory and radar scene structure simultaneously through differentiable radar measurements, eliminating heuristic scan matching and improving robustness in degenerate sensing conditions.

\item \textbf{System implementation and evaluation.}
We implement DiffRadar on commodity FMCW radar hardware and evaluate it on both the public benchmark and a Radar Degeneracy Stress-Test (RDST) suite for radar-specific failure modes, including corridor degeneracy, speed transitions, dynamic clutter, and long-horizon drift. 
DiffRadar achieves up to a $6\times$ reduction in benchmark trajectory error and over $20\times$ improvement under feature-poor corridor motion, while maintaining real-time performance at 70 FPS and more than doubling map consistency.
\end{itemize}

\section{Radar Sensing and Representation}
{\textbf{Radar Sensing and FMCW Fundamentals}}
Millimeter-wave (mmWave) radars are increasingly accessible due to advances in single-chip frequency-modulated continuous wave (FMCW) technology \cite{klarie2015single,ti_iwr1843,ti_fmcw_fundamentals}. 
These sensors transmit chirps and measure reflected echoes to estimate the range, angle, and Doppler shift of surrounding reflectors.
For a chirp with slope $S$ and round-trip delay $\tau = 2r/c$, the beat frequency $f_b = S\tau$ determines the target range $r$, while radial motion produces a Doppler shift $f_D = 2v/\lambda$. 
A key property of FMCW sensing is its \textit{anisotropic spatial resolution}: radar achieves high precision in range but much coarser angular resolution due to the limited antenna aperture \cite{ti_iwr1843}. 

{\textbf{Limitations of Discrete Radar Representations}}
Conventional radar SLAM systems represent the environment using occupancy grids or RA heatmaps \cite{hess2016real,prabhakara2023high,cen2019radar,hong2021radar,sie2024radarize}. 
While convenient for visualization and processing, such discretization poorly reflects the physical structure of radar sensing. 
Radar measurements exhibit correlated uncertainties introduced by beamforming and antenna sidelobes \cite{ti_fmcw_fundamentals}, yet grid representations treat cells independently and reduce the continuous signal formation process to per-cell intensities. 
Localization is therefore typically performed through scan matching between discrete frames, which can become unstable in feature-sparse or repetitive environments \cite{sie2024radarize}. 
Although DA cues can improve odometry, existing approaches still rely on discretized radar heatmaps and modular estimators that separate translation and rotation inference \cite{sie2024radarize,lu2020milliego}.

{\textbf{Towards Differentiable Radar Modeling}}
Recent advances in visual scene representation demonstrate the potential of differentiable modeling. 
In particular, 3D Gaussian Splatting (3DGS) represents scenes using anisotropic Gaussian primitives that can be rendered differentiably through optical alpha-compositing \cite{kerbl20233d}, enabling efficient optimization for tasks such as view synthesis and visual SLAM \cite{yugay2023gaussian,keetha2024splatam,zhang2026rf}. 

Recent work has explored neural scene representations for radar sensing, including neural radar fields and radar-aware Gaussian representations \cite{rafidashti2025neuradar,lei2024sar,zhang2026rf}. 
However, radar signal formation differs fundamentally from optical rendering. Radar echoes arise from additive reflections governed by path loss and directional backscatter rather than optical occlusion, meaning that measurements may combine contributions from multiple scatterers along the same beam. Consequently, optical rendering models such as 3DGS are inconsistent with radar sensing physics.

This observation motivates retaining the continuous Gaussian representation of 3DGS while replacing its optical renderer with a radar-consistent forward model aligned with FMCW sensing and antenna array physics. 
Such a formulation enables analytic gradients through both geometric and Doppler domains, allowing mapping and motion estimation to be optimized jointly within a differentiable framework. This design forms the basis of our DiffRadar system.

\begin{figure*}[t]
\centering
\includegraphics[width=0.9\textwidth]{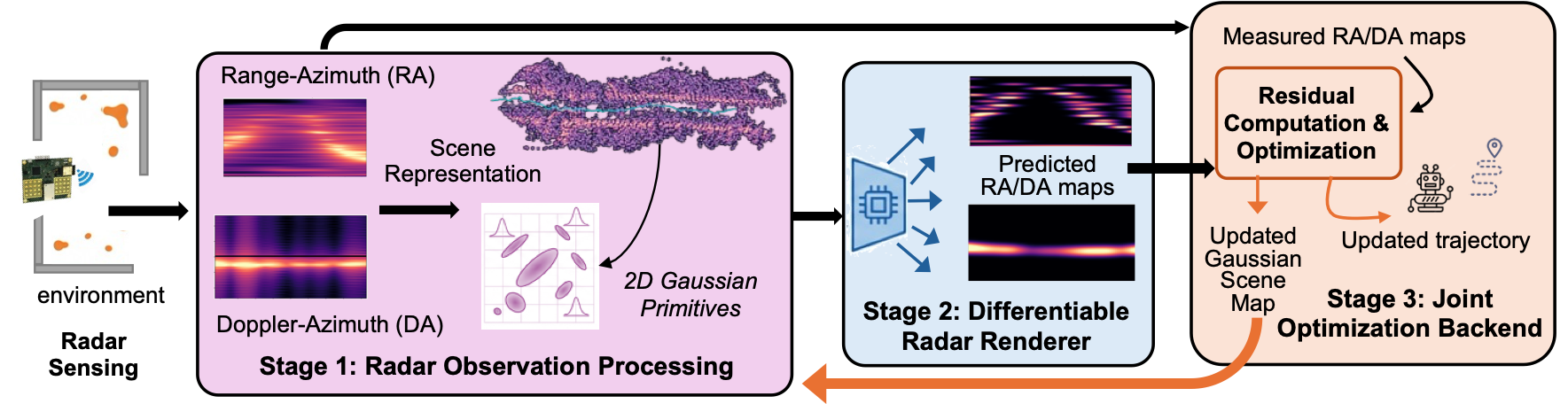}
\caption{{DiffRadar pipeline.}
Radar observations are processed into RA and DA maps, which initialize Gaussian primitives representing the scene.
A differentiable renderer projects the Gaussian scene into radar measurement space, and residuals between rendered and observed measures drive joint optimization of the scene and trajectory.}\label{fig:overview}
\Description{System pipeline of DiffRadar. Radar sensing produces range-azimuth and Doppler-azimuth maps, which initialize 2D Gaussian primitives. A differentiable radar renderer predicts RA and DA measurements from the Gaussian scene, and residuals between predicted and measured radar maps are used by a joint optimization backend to update both the Gaussian map and the trajectory.}

\end{figure*}

\section{DiffRadar Overview}
DiffRadar reformulates radar SLAM as a differentiable optimization problem grounded in radar sensing physics. 
Instead of separating signal processing, motion estimation, and mapping into independent stages, DiffRadar integrates radar measurements, scene representation, and trajectory estimation within a unified optimization framework. 
This design enables scene structure and robot pose to be optimized jointly from radar observations.

Designing such a system introduces three key challenges.
First, radar observations exhibit anisotropic uncertainty in range and angle and are naturally expressed in polar coordinates, requiring a continuous representation compatible with gradient-based optimization.
Second, radar measurements depend not only on geometry but also on sensing physics, including directional backscatter, Doppler shifts, and sensor resolution, which must be captured in the scene representation.
Third, motion estimation and mapping must be tightly coupled so that geometric and Doppler constraints can be exploited simultaneously during optimization.

\paragraph{\textbf{System Overview}}
Figure~\ref{fig:overview} shows DiffRadar architecture, which forms a differentiable radar SLAM pipeline with three tightly coupled stages. 
Sections~\ref{sec:representation} to~\ref{sec:optimization} describe the scene representation and optimization procedure in detail.

\textbf{Stage 1: Radar Observation Processing.}
Raw radar I/Q measurements are processed through standard FMCW signal processing to produce RA and DA radar maps. 
Gaussian primitives are initialized from CFAR detections to provide an initial scene representation.

\textbf{Stage 2: Differentiable Radar Rendering.}
The Gaussian scene representation is projected into radar measurement space using a differentiable radar renderer.
The renderer models radar signal formation using an additive reflection model consistent with FMCW sensing and antenna beam patterns.
It produces synthetic RA and DA radar maps whose residuals with respect to observed measurements provide gradients for jointly optimizing scene parameters and trajectory.

\textbf{Stage 3: Joint Optimization Backend.}
A unified optimization backend minimizes residuals between rendered and observed radar measurements to refine both scene structure and trajectory.
Range–azimuth observations constrain geometry and orientation, while Doppler measurements provide complementary constraints on translation and velocity.

\section{Physics-Aware Radar Representation}\label{sec:representation}

DiffRadar represents the environment as a continuous radar scene field parameterized by Gaussian primitives that encode both spatial geometry and radar scattering behavior. 
This representation allows radar observations to be rendered differentiably, enabling gradient-based refinement of both the scene structure and robot trajectory. 
This section describes the physics-grounded parameterization of these primitives.

\subsection{Gaussian Primitive Model}
\begin{figure}[t]
  \centering
  \includegraphics[width=0.45\textwidth]{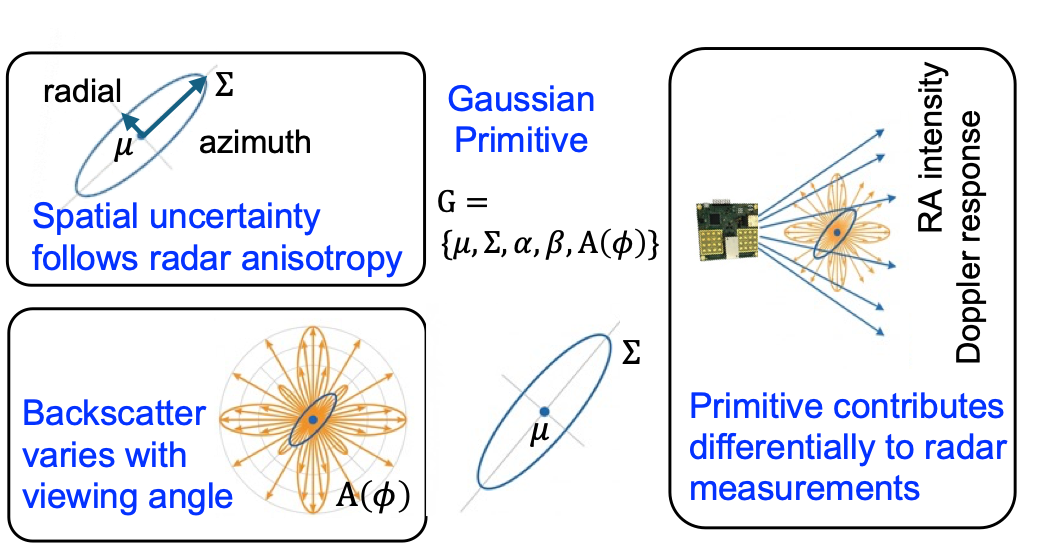}
\caption{Physics-aware Gaussian primitive in DiffRadar. 
Each scatterer is modeled as a Gaussian capturing spatial anisotropy, directional backscatter, and contributions to RA and DA measurements.}
  \label{fig:gaussian_components}
  \Description{Illustration of a physics-aware Gaussian primitive used to model a radar scatterer. The primitive has anisotropic spatial uncertainty aligned with radar range and azimuth resolution, angle-dependent backscatter, and a contribution to both range-azimuth intensity and Doppler-azimuth response.}

\end{figure}

Figure~\ref{fig:gaussian_components} illustrates the Gaussian primitive used in DiffRadar.  
The environment is represented as a collection of primitives
\begin{equation}
G_i = \{\mu_i, \Sigma_i, \beta_i, A_i(\phi), \alpha_i\}, \label{eq:primitive}
\end{equation}
where $\mu_i$ denotes the spatial mean of primitive $i$ in the global map frame and $\Sigma_i$ its covariance capturing spatial uncertainty. 
The scalar $\beta_i$ represents the reflectivity of the primitive, $A_i(\phi)$ models the directional dependence of radar backscatter with respect to azimuth angle $\phi$, and $\alpha_i \in [0,1]$ is a reliability weight derived from the signal-to-noise ratio (SNR).

Each primitive jointly captures spatial geometry, radar sensing anisotropy, and viewpoint-dependent scattering behavior. 
Representing radar scatterers as Gaussian primitives provides a continuous approximation of sparse radar measurements, enabling stable gradients for both pose and map parameters. 
This formulation parallels the use of Gaussian primitives in 3D Gaussian Splatting for differentiable scene modeling in vision-based systems \cite{matsuki2024gaussian}.

\subsection{Radar Anisotropy Modeling}
MmWave radars provide fine spatial resolution along range but significantly coarser separation across azimuth due to the limited antenna aperture. 
To capture this anisotropy, DiffRadar models each primitive's spatial uncertainty using a rotation-scale decomposition
\begin{align}
\Sigma_i = R_i S_i^2 R_i^\top , \label{eq:cov_decomp}
\end{align}
where $R_i$ rotates the local range-cross-range coordinate system into the global map frame and 
$S_i = \mathrm{diag}(\sigma_r, \sigma_\theta r_i)$ encodes the anisotropic spatial scales along the radial and cross-range directions. 
Here $r_i$ denotes the range from the radar to the primitive, while $\sigma_r$ and $\sigma_\theta$ correspond to the radar's intrinsic sensing resolution. 
We have $\sigma_r = \frac{c}{2B}$ and $\sigma_\theta = \frac{\lambda}{D}$, where $c$ is the speed of light, $B$ the FMCW sweep bandwidth, $\lambda$ the wavelength, and $D$ the effective antenna aperture.
This construction produces primitives with narrow radial extent and cross-range uncertainty that grows with distance, consistent with the sensing geometry of FMCW radar.

\subsection{Directional Backscatter}
Radar reflections depend strongly on viewing direction due to surface geometry and antenna beam patterns. 
To model this effect, DiffRadar represents the azimuth-dependent reflectivity of each primitive using a low-order harmonic expansion
$A_i(\phi)=\sum_{k=-K}^{K} a_k e^{jk\phi}$,
where coefficients $a_k$ capture the directional scattering profile of the primitive. 
In practice, a small order $K \le 2$ is sufficient to represent the smooth angular variations typically observed in mmWave radar returns while avoiding overfitting to noise. 
The final directional reflectivity of primitive $i$ is
\begin{align}
R_i(\phi) = \beta_i A_i(\phi),
\end{align}
which provides a continuous approximation of viewpoint-dependent scattering. 
As reflectivity varies with observation angle, changes in pose produce predictable changes in rendered measurements, allowing directional scattering to contribute informative gradients for mapping and localization.

Together, $(\Sigma_i, A_i(\phi), \beta_i)$ provide a physics-aligned parameterization linking spatial geometry with radiometric behavior, enabling coherent optimization of scene structure and sensor motion.

\section{Differentiable Radar Rendering}\label{sec:renderer}
Given the Gaussian scene representation, DiffRadar predicts radar observations using a physics-consistent differentiable radar renderer. 
Unlike optical rendering in 3D Gaussian Splatting, radar returns accumulate additively from multiple scatterers along each beam. 
The renderer maps scene primitives and robot pose to RA and DA measurements, enabling gradient-based refinement of both scene parameters and trajectory.

\subsection{Range-Azimuth (RA) Rendering}
Radar signal propagation is additive: the received power at a beam $(\phi,r)$ equals the sum of contributions from all illuminated scatterers. 
DiffRadar models this process using a differentiable renderer that projects Gaussian primitives into the radar's polar observation space:
\begin{align}
I(\phi,r,t)=
\sum_{i} \mathcal{A}_t(i)\,
\alpha_i\,\beta_i\,A_i(\phi)\,G_i(r;\mu_i,\mathbf{T}_t),
\label{eq:render}
\end{align}
where $G_i(r;\mu_i,\mathbf{T}_t)$ denotes a Gaussian kernel centered at the projected range of primitive $i$ after transforming its mean $\mu_i$ by the current pose $\mathbf{T}_t$. 
The association mask $\mathcal{A}_t(i)$ (computed via the visibility mechanism in Section~\ref{sec:optimization}) ensures that gradients propagate only via physically plausible scatterers, which is important for radar measurements that may include spurious reflections.
The factor $\alpha_i$ down-weights unreliable or transient scatterers and is initialized from the normalized SNR. 
Because each primitive is transformed by the current pose estimate, the rendered intensity depends smoothly on both scene parameters and trajectory variables, enabling stable gradient-based optimization.

\subsection{Doppler Rendering}
Unlike reflectivity or spatial extent, Doppler arises from the relative motion between the radar and the scene. 
For a primitive at position $\mu_i$, the Doppler shift induced by ego-motion $\mathbf{v}(t)$ is
$f_{D,i}(t)$=$\frac{2}{\lambda}\hat{\mathbf r}_i(t)^\top$ $\mathbf{v}(t)$,
where $\hat{\mathbf r}_i(t)$ is the unit vector from the radar to the primitive and $\lambda$ is the radar wavelength. 

Radar systems estimate Doppler per azimuth beam rather than per individual scatterer, since returns from multiple scatterers are aggregated within each beam. 
DiffRadar therefore predicts the Doppler-azimuth (DA) profile by aggregating contributions from all primitives overlapping each beam:
\begin{align}
\hat f_D(\phi,t)=\sum_{i} \mathcal{A}_t(i)\, w_i(\phi,t)\, f_{D,i}(t),
\label{eq:dopp_render}
\end{align}
where the weights $w_i(\phi,t)$ reflect spatial proximity and radiometric strength. 
In practice, we define $w_i(\phi,t)$ $\propto$ $\alpha_i \beta_i A_i(\phi)$ $G_i(r;\mu_i,\mathbf{T}_t)$,
and normalize across all contributing primitives. 
This formulation ties Doppler prediction to the same physical factors governing RA intensity while keeping computation lightweight.
Doppler residuals provide strong translation-sensitive gradients: modifying $\mu_i$ or $\mathbf{v}(t)$ changes $f_{D,i}(t)$ in directionally informative ways. 
These cues complement the geometric gradients derived from intensity rendering and enable accurate motion estimation even in environments with weak structural features.

\begin{figure}[t]
\centering
\includegraphics[width=\linewidth]{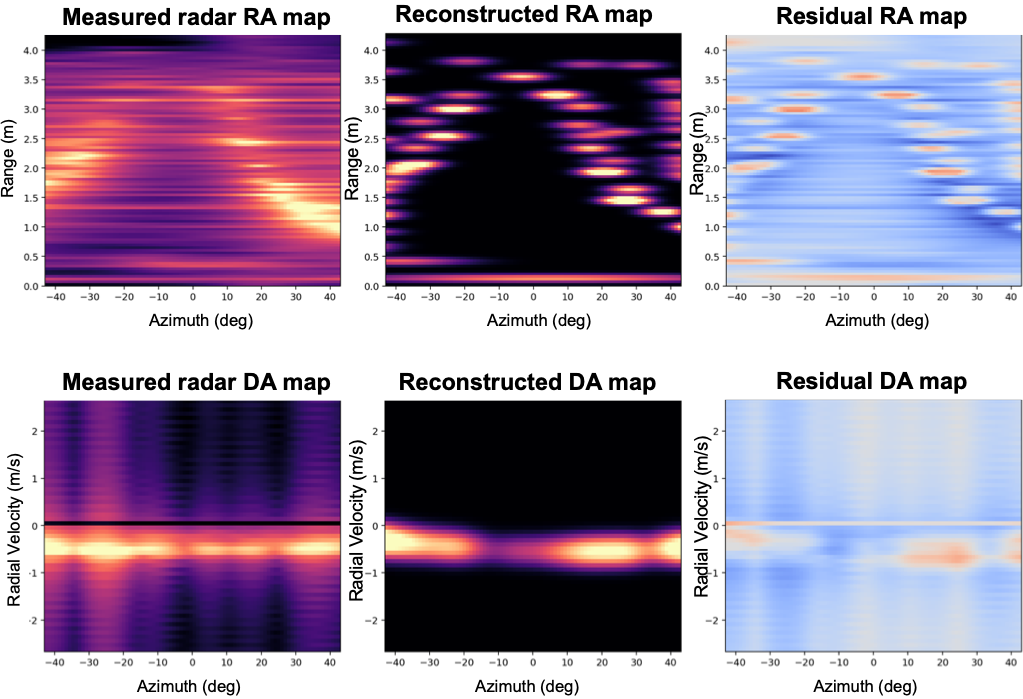}
\caption{Measured and rendered radar observations used for optimization.
Left: measured radar maps.
Middle: maps rendered from the Gaussian scene representation.
Right: residual maps used in the optimization objective.
RA intensity (top), DA responses (bottom).}
\label{fig:render_residual}
\Description{Side-by-side comparison of measured radar observations, radar observations rendered from the Gaussian scene representation, and residual maps used for optimization. The top row shows range-azimuth intensity maps and the bottom row shows Doppler-azimuth responses, highlighting how optimization is driven by differences between measured and predicted radar signals.}

\end{figure}

\subsection{Unified Optimization Objective}
DiffRadar jointly optimizes robot pose and Gaussian primitive parameters by minimizing residuals between rendered and measured radar observations:
\begin{align}
\mathcal{L} (\mathbf{T}_t, \{G_i\}) =
\lambda_{RA}\|I_{\text{rend}} - I_{\text{meas}}\|_2^2 +
\lambda_{DA}\|\hat f_D - f_{D,\text{meas}}\|_2^2 .
\label{eq:loss_total}
\end{align}

Figure~\ref{fig:render_residual} illustrates this principle: the differentiable radar renderer generates synthetic radar maps whose residuals with respect to measured observations provide gradients for joint pose and scene refinement. 
The RA term enforces geometric consistency between rendered and measured radar intensity, adjusting primitive geometry and pose parameters. 
The Doppler term couples velocity, primitive geometry, and pose through the kinematic projection in Eq.~\eqref{eq:dopp_render}, providing strong translation-sensitive gradients.

Together, Eqs.~\eqref{eq:render} and \eqref{eq:dopp_render} link scene geometry, radar scattering, and motion through a unified differentiable objective. Section~\ref{sec:optimization} describes how radar-side associations combine with these gradients to enable real-time SLAM.
\section{Online SLAM System}\label{sec:optimization}
While Sections~\ref{sec:representation} and~\ref{sec:renderer} define the radar scene representation and differentiable rendering model, this section describes how they are integrated into an online SLAM backend. 
Radar observations are sparse, anisotropic, and Doppler-coupled, making direct optimization challenging. 
DiffRadar addresses this by combining radar-conditioned visibility with differentiable optimization to jointly refine the robot trajectory and Gaussian scene map.
Algorithm~\ref{alg:diffradar} summarizes the online pose--map optimization procedure.

\begin{algorithm}[t]
\caption{DiffRadar Online SLAM}
\label{alg:diffradar}
\begin{algorithmic}[1]
\Require Incoming radar frames $\{I_t^{RA}, I_t^{DA}\}$
\Ensure Estimated poses $\{\mathbf{T}_t\}$ and Gaussian map $\mathcal{G}$

\State \textbf{Initialization:}
\State Estimate rough initial poses using lightweight radar odometry
\State Initialize Gaussian primitives from CFAR detections
\State $\mathcal{G} \gets \{\mu_i,\Sigma_i,\alpha_i,\beta_i,A_i(\phi)\}$

\For{each incoming radar frame $t$}

    \State Extract detections and update candidate primitives
    \State Compute association set $\mathcal{A}_t$ using VDRF

    \State Render predicted RA/DA measurements $\hat I_t^{RA}, \hat f_{D,t}$

    \State Compute residual loss $\mathcal{L}$ using Eq.~\eqref{eq:loss_total}

    \State Update pose parameters $(\mathbf{T}_t,\mathbf{v}(t))$ and Gaussian 
    \Statex \hspace{1em}  primitive parameters $(\mu_i,\Sigma_i,\beta_i,A_i)$

    \State Apply primitive lifecycle management: insert 
    \Statex \hspace{1em}  supported primitives and prune stale ones

\EndFor

\State \Return optimized trajectory $\{\mathbf{T}_t\}$ and map $\mathcal{G}$
\end{algorithmic}
\end{algorithm}

\subsection{Primitive Lifecycle}
The radar front-end converts radar detections into Gaussian primitives that anchor the differentiable SLAM optimization. 
Because mmWave radar measurements contain noise, sidelobes, and multipath artifacts, DiffRadar maintains a stable scene representation through primitive initialization, visibility-conditioned association, and lifecycle management.

\textbf{Initialization and lifecycle management.}
CFAR (constant false alarm rate) detections in each RA frame instantiate Gaussian primitives using the parameterization defined in Section~\ref{sec:representation}.
Primitive means and covariances inherit the radar's anisotropic sensing resolution, while reflectivity and reliability are initialized from measured power and SNR. 
Directional scattering coefficients are initialized without prior structure and refined during optimization. 
During online SLAM, primitives are activated only after repeated observation support, stale primitives are removed when no longer visible, and redundant scatterers are merged to maintain a compact and stable scene representation.

\textbf{Visibility-conditioned association (VDRF).}
Given the current radar frame, DiffRadar determines which primitives can physically contribute to the measurement using a visibility function $\mathcal{V}_t(i)\in\{0,1\}$ that extends visibility concepts from visual 3DGS to radar sensing. 
A primitive is considered visible when its geometry, angular scattering, and Doppler-consistent motion jointly produce sufficient likelihood under the radar forward model:
\[
\mathcal{V}_t(i)=
\begin{cases}
1, & p(z_t \mid G_i,\mathbf{T}_t) > \tau,\\
0, & \text{otherwise}.
\end{cases}
\]

The likelihood is factorized as
\[
p(z_t \mid G_i,\mathbf{T}_t)
=
p(r_t \mid \mu_i,\Sigma_i)\,
p(\phi_t \mid A_i(\phi))\,
p(f_{D,t} \mid f_{D,i}(t)),
\]
where the three terms capture range consistency, angular scattering support, and Doppler-consistent motion for each detection $z_t=(r_t,\phi_t,f_{D,t})$. 
The resulting association set $\mathcal{A}_t=\{i\mid\mathcal{V}_t(i)=1\}$ determines which primitives participate in the differentiable rendering in Eq.~\eqref{eq:render}. 
Unlike RGB-based 3DGS where visibility emerges from ray sampling, radar requires explicit physics-conditioned visibility to prevent gradients from propagating through spurious scatterers.

\subsection{Joint Pose-Map Optimization}
Given the association set $\mathcal{A}_t$, DiffRadar jointly refines the radar scene representation and robot trajectory by minimizing the residual objective in Eq.~\eqref{eq:loss_total}. 
Gradients derived from RA intensity and Doppler measurements update complementary aspects of the scene and motion parameters.

RA intensity residuals primarily constrain the spatial structure of the scene. 
These gradients update the primitive means $\mu_i$ and covariances $\Sigma_i$, shifting primitives toward RA configurations that better explain observed radar returns while preserving radar-consistent anisotropy through the covariance structure in Eq.~\eqref{eq:cov_decomp}. 
The same residuals also refine the radiometric parameters of each primitive, including reflectivity $\beta_i$ and the directional scattering function $A_i(\phi)$, allowing each primitive to learn a continuous azimuth-dependent backscatter pattern consistent with the measurements.

Doppler residuals provide motion-sensitive constraints. 
Because Doppler depends on both primitive geometry and ego-motion $\mathbf{v}(t)$, minimizing Doppler error refines the velocity estimate while simultaneously adjusting primitive positions in translation-sensitive directions. 
These kinematic gradients complement the geometric constraints from RA measurements and significantly improve pose estimation in environments with sparse structural features.

\begin{figure}[t]
\centering
\includegraphics[width=0.8\linewidth]{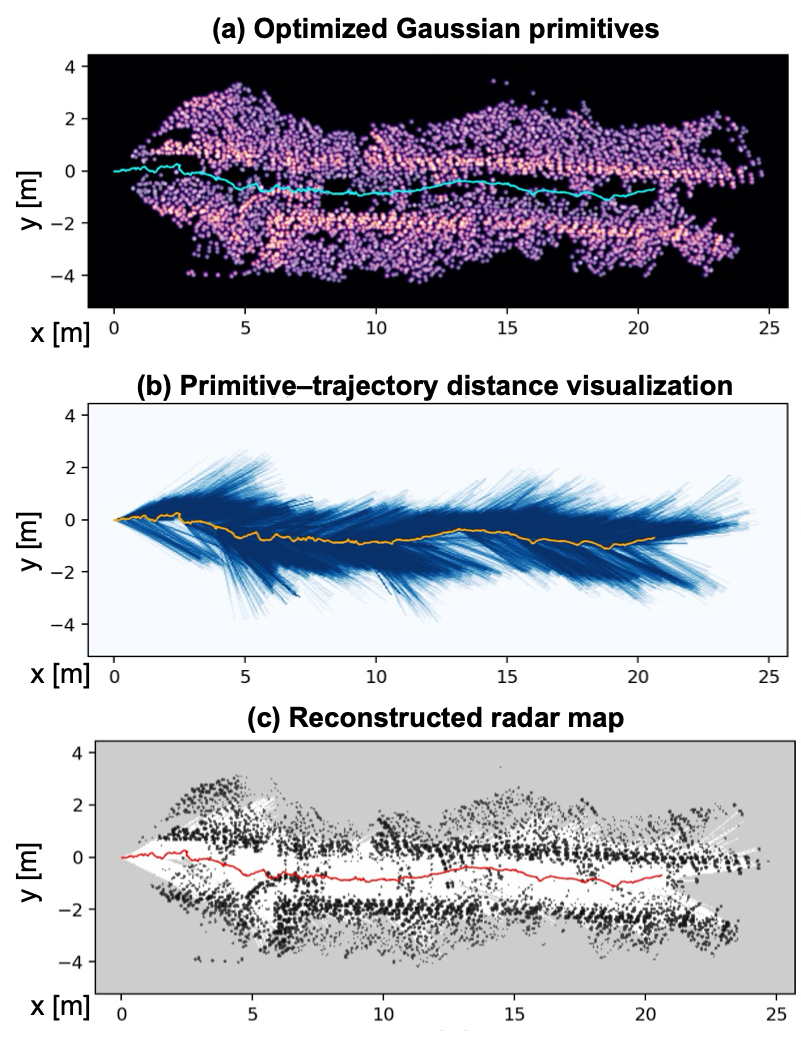}
\caption{{Gaussian scene map constructed by DiffRadar.}
Top: optimized Gaussian primitives (pink) representing radar scatterers.
Middle: distances between trajectory states and nearby primitives.
Bottom: reconstructed radar map along the estimated trajectory.}
\label{fig:gaussian_map}
\Description{Visualization of the Gaussian scene map estimated by DiffRadar. The top panel shows optimized Gaussian primitives representing radar scatterers, the middle panel shows distances between trajectory states and nearby primitives, and the bottom panel shows the reconstructed radar map along the estimated trajectory.}
\end{figure}

Optimization proceeds through alternating pose and map refinement. 
With the map fixed, trajectory parameters $(\mathbf{T}_t,\mathbf{v}(t))$ are updated to explain the current radar frame. 
With the pose fixed, gradients update primitive parameters including geometry $(\mu_i,\Sigma_i)$, reflectivity $\beta_i$, and directional scattering $A_i(\phi)$. 
After each update, primitive covariances are reprojected to maintain the radar-consistent anisotropic structure defined in Eq.~\eqref{eq:cov_decomp}.
Figure~\ref{fig:gaussian_map} illustrates the Gaussian scene map produced during SLAM. 
The optimized primitives form a structured radar representation that supports differentiable rendering and consistent pose--map refinement along the estimated trajectory. 
The resulting Gaussian scene field enables consistent joint estimation of trajectory and radar map structure.
\section{Evaluation}
\label{sec:eval}
We evaluate DiffRadar across four aspects:
(i) SLAM accuracy,
(ii) robustness under radar degeneracy,
(iii) map fidelity, and
(iv) runtime efficiency.
Experiments are conducted on two complementary datasets.
We first evaluate trajectory accuracy on the public Radarize benchmark to enable direct comparison with prior radar SLAM systems.
Second, we introduce a controlled degeneracy stress-test suite that isolates radar-specific failure regimes rarely captured in existing SLAM benchmarks, including corridor degeneracy, motion regime transitions, dynamic clutter, and long-horizon loop closures.
Across these experiments, we evaluate trajectory accuracy, robustness under degeneracy and dynamic clutter, map consistency, and runtime performance.

\subsection{Experimental Setup}
\label{sec:eval_setup}

\begin{figure}[t]
\centering
\includegraphics[width=0.8\linewidth]{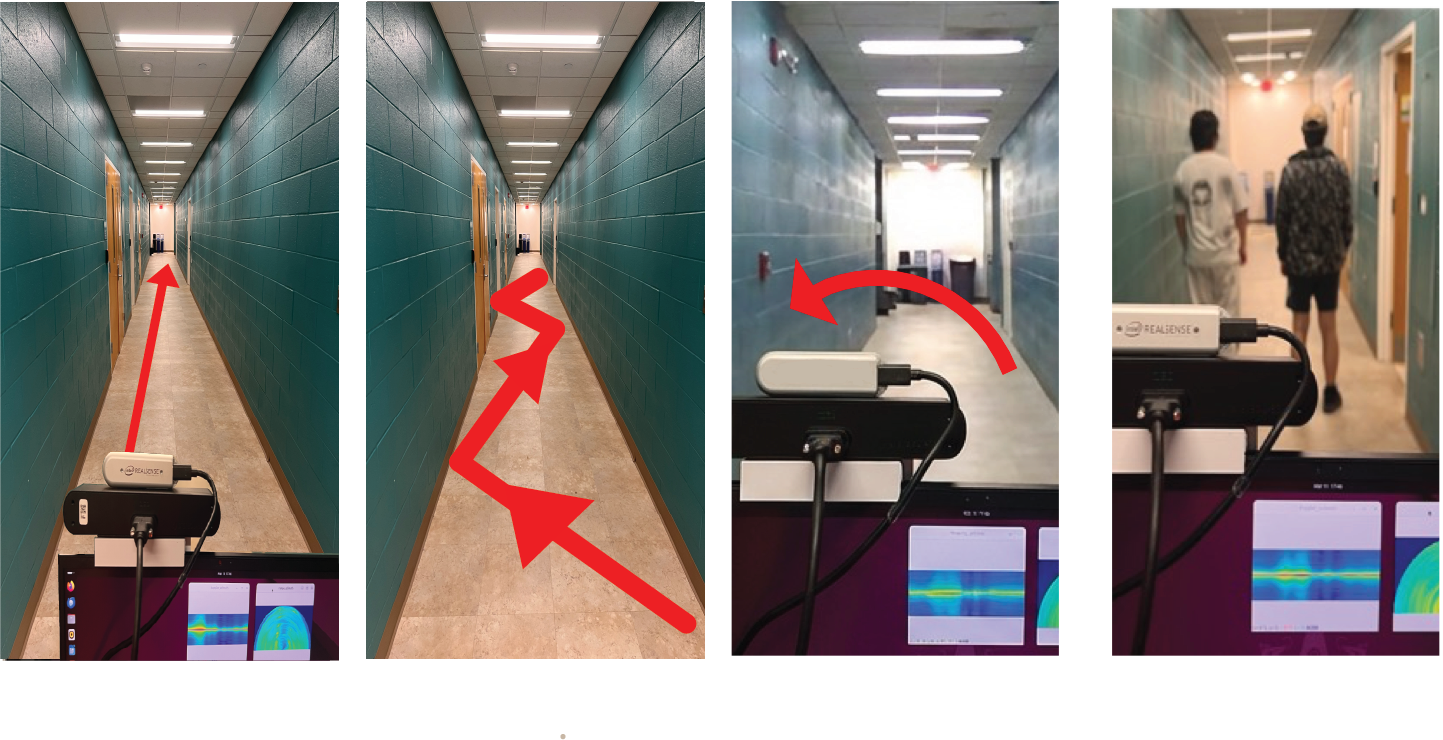}
\caption{Representative environments in the RDST stress-test suite.
From left to right: a long feature-poor corridor, zig-zag motion with lateral excitation,
in-place rotation sequences, and dynamic clutter with moving pedestrians.
Additional cases, including scripted speed transitions and long loop closures ($>$100\,m),
are evaluated but not visualized here.}
\label{fig:RDST_scenarios}
\Description{Example environments and motion regimes from the RDST stress-test suite. The figure shows representative cases including a long feature-poor corridor, zig-zag motion with lateral excitation, in-place rotation, and dynamic clutter with moving pedestrians, illustrating challenging conditions used to evaluate radar SLAM robustness.}

\end{figure}

{\textbf{Datasets.}}
We evaluate on two datasets.
\textbf{(1) Radarize benchmark.}
For direct comparison with prior radar SLAM systems, we use the public Radarize dataset \cite{sie2024radarize}, which provides synchronized RA and DA measurements and is widely used for radar-only odometry and SLAM evaluation.
We follow the original dataset splits. 
Pseudo ground-truth trajectories are obtained from Intel RealSense T265 visual-inertial odometry; evaluation uses planar $(x,y,\psi)$ components after SE(2) alignment.
\textbf{(2) Radar Degeneracy Stress-Test (RDST) Suite.}
To evaluate radar SLAM under controlled failure regimes, we construct a RDST suite consisting of 52 trajectories collected across two indoor sites (approximately 1.8\,km total) using handheld and cart-mounted platforms.
These trajectories introduce motion and sensing conditions designed to expose challenging radar SLAM regimes, including 
(i) zig-zag motion with lateral excitation,
(ii) pure in-place rotations,
(iii) dynamic clutter with moving humans,
(iv) scripted speed transitions (slow$\rightarrow$fast$\rightarrow$slow), and
(v) long loop closures exceeding 100\,m.
Figure~\ref{fig:RDST_scenarios} shows representative scenarios.
Ground-truth trajectories are obtained using stereo/RGB-D visual-inertial SLAM and aligned to the radar frame via SE(2).

{\textbf{Hardware and radar processing.}}
Experiments use commodity 77\,GHz FMCW radar platforms.
For the Radarize benchmark, we use the TI IWR1843 radar config employed in the original dataset collection \cite{sie2024radarize}.
For the RDST dataset, we collect data using a TI AWR1843BOOST radar with a different antenna configuration and mounting setup.

For each radar frame, we compute (i) a RA intensity map and (ii) a DA map.
To ensure fair comparison on the public benchmark, we follow Radarize-style preprocessing settings, including comparable maximum range, range resolution, angular binning, and frame rate \cite{sie2024radarize}.
DiffRadar consumes the same RA/DA inputs but refines pose and map estimates through differentiable radar rendering residuals instead of discrete scan matching. 
Unless otherwise stated, hyperparameters remain fixed across datasets.

{\textbf{Implementation details.}}
DiffRadar is implemented in PyTorch as an online fixed-lag SLAM optimizer over a dynamic Gaussian map. Each radar scatterer is represented by a Gaussian primitive with learnable position, anisotropic covariance, and reflectivity, initialized from RA detections and jointly refined with platform pose. 
We optimize a sliding window of 8--12 frames using alternating pose and map updates with Adam and a robust Huber loss; Doppler consistency provides a velocity prior, while short-term motion regularization improves temporal stability. 
To keep the online map bounded, new primitives are confirmed only after persistent multi-view support and low-confidence or stale primitives are pruned during execution. 
Unless otherwise stated, the same optimization and map-management settings are used across all datasets and experiments; after stabilization, the active map typically contains approximately $100$ Gaussian primitives at any given time, although over the full sequence the system may instantiate thousands cumulatively as old primitives are pruned or merged.

\textbf{Metrics.}
We report \textit{Absolute Trajectory Error (ATE)} and \textit{Relative Error (RE)} for both translation and yaw.
ATE is computed after global SE(2) alignment to the reference trajectory, while RE is computed on fixed-length sub-trajectories.
Loop drift is measured at loop closure and normalized by trajectory length (m per 100\,m).
For the RDST stress tests, we additionally evaluate robustness under controlled radar failure modes.
\textit{Corridor stall rate} measures the fraction of frames where the estimated translation magnitude falls below $\epsilon$ while ground-truth motion exceeds $\epsilon$.
\textit{Pure-rotation yaw error} measures final yaw deviation after scripted in-place rotations.
\textit{Speed-change error} captures transient pose or velocity errors following scripted slow$\leftrightarrow$fast transitions.
\textit{Dynamic-clutter robustness} measures ATE and RE degradation as the number of moving humans increases (0, 2, 4).

To evaluate map quality, we report \textit{Map consistency (MapCons)}, defined as the overlap between projected radar maps and occupancy derived from depth sensing, and \textit{Vertical artifact rate (V-Artifact)}, the fraction of reconstructed primitives violating inferred floor or ceiling constraints.

Finally, we report system efficiency including steady-state runtime, stage-wise latency, end-to-end frame rate (FPS), and scalability with respect to the number of active Gaussian primitives and the fixed-lag optimization window.

{\textbf{Baselines.}}
We compare DiffRadar against three representative baselines:
(i) \textit{Radarize}, a radar-only SLAM system that combines Doppler-based translation estimation with correlation or learning-based rotation estimation on RA/DA heatmaps \cite{sie2024radarize};
(ii) \textit{milliEgo}, a radar–inertial odometry system that fuses mmWave radar with IMU measurements \cite{lu2020milliego};
(iii) \textit{RNIN}, a neural inertial navigation method based on recurrent modeling of IMU sequences \cite{chen2021rnin}.
Where feasible, we re-run baselines under identical preprocessing settings; otherwise we report published results and mark them with~$\dagger$.



\subsection{Benchmark Evaluation on Radarize} \label{sec:results_radarize}

\begin{table}[!tbp]
\centering
\caption{\textbf{End-to-end SLAM trajectory accuracy (public benchmark)}.
$\dagger$ indicates reported values.}
\label{tab:slam_traj}
\small
\setlength{\tabcolsep}{6pt}
\renewcommand{\arraystretch}{1.15}
\begin{tabular}{lrrrr}
\toprule
Method & ATE$_t$\down (m) & ATE$_\psi$\down (rad) & RE$_t$\down (m) & RE$_\psi$\down (rad) \\
\midrule
RNIN$^\dagger$     & 3.299  & 0.541  & 0.020  & 0.522  \\
milliEgo$^\dagger$ & 3.900  & 0.808  & 0.022  & 0.780  \\
Radarize$^\dagger$ & 0.606  & 0.116  & \best{0.017} & 0.113 \\
\midrule
\textbf{DiffRadar} & \best{0.103} & \best{0.084} & 0.025 & \best{0.034} \\
\bottomrule
\end{tabular}
\end{table}

\textbf{End-to-End SLAM Accuracy.}
We first evaluate DiffRadar on the public Radarize benchmark using the same evaluation protocol as prior work.
Table~\ref{tab:slam_traj} summarizes trajectory accuracy across all sequences. 
Overall, DiffRadar achieves the best global trajectory accuracy among all evaluated methods on the Radarize benchmark. 
In particular, DiffRadar reduces translational ATE from $0.606$m (Radarize) to $0.103$m, corresponding to roughly a $6\times$ improvement, while also achieving the lowest rotational error among all methods.
These improvements arise from jointly optimizing RA and DA measurements within a physics-aware differentiable radar model.
Doppler observations provide strong translation-sensitive constraints, while RA rendering residuals stabilize orientation and enforce map consistency.
Optimizing both signals within a unified pose--map framework enables DiffRadar to maintain stable trajectory estimates even in geometrically ambiguous environments such as long corridors.

DiffRadar exhibits slightly higher relative translation error (RE$_t$) than Radarize.
This difference reflects the design trade-off between local odometry accuracy and global trajectory consistency.
Radarize estimates instantaneous translation directly from Doppler measurements, which yields strong short-term motion estimates.
In contrast, DiffRadar jointly optimizes pose and scene parameters, prioritizing globally consistent trajectories and significantly reducing long-horizon drift.

\begin{figure}[t]
\centering
\includegraphics[width=0.9\linewidth]{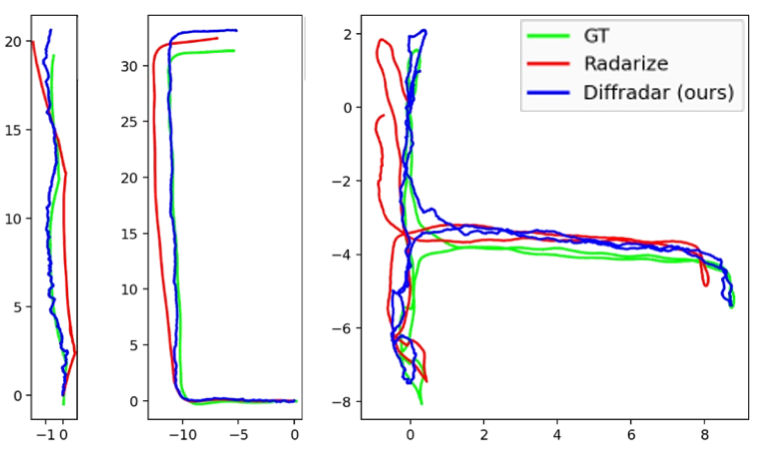}
\caption{Top-down trajectory comparison on three Radarize sequences (units in meters).
Columns show a narrow corridor sequence, a long L-shaped corridor trajectory, and a hallway intersection with multiple turns.
Radarize (blue) accumulates drift in corridor segments and around turns, while DiffRadar (red) closely follows the reference trajectory (black).}
\label{fig:traj_compare}
\Description{Top-down trajectory plots for three representative benchmark sequences: a narrow corridor, a long L-shaped corridor, and a hallway intersection with multiple turns. In each case, DiffRadar stays closer to the reference trajectory, while Radarize shows larger drift in straight corridors and around turns.}

\end{figure}

\begin{figure}[t]
\centering
\includegraphics[width=0.9\linewidth]{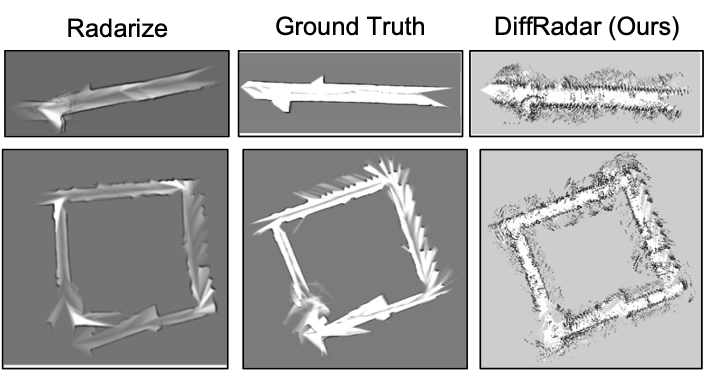}
\caption{Radar map reconstruction comparison.
Columns show Radarize, ground truth, and DiffRadar.
Top: long corridor sequence with limited lateral structure.
Bottom: square-loop trajectory requiring consistent loop closure.
DiffRadar produces sharper and more geometrically consistent radar maps.}
\label{fig:map_compare}
\Description{Map reconstruction comparison for two scenarios, with columns for Radarize, ground truth, and DiffRadar. The top row shows a long corridor sequence and the bottom row shows a square-loop trajectory. Radarize produces blur, duplication, and geometric distortion, while DiffRadar reconstructs sharper and more consistent map structure.}

\end{figure}

\textbf{Qualitative Results.}
Figure~\ref{fig:traj_compare} compares estimated trajectories for three representative Radarize sequences.
The scenes include a feature-poor corridor, a long corridor with a sharp turn, and a hallway intersection with multiple turns.
Such environments are challenging for radar SLAM due to limited lateral structure and motion ambiguity.
Radarize accumulates noticeable drift along long corridors and near turns.
In contrast, DiffRadar remains closely aligned with the reference trajectory, indicating that joint pose--map optimization improves motion estimation stability under challenging radar conditions.

Figure~\ref{fig:map_compare} compares reconstructed radar maps for two representative scenarios. 
In the long corridor sequence, the Radarize baseline produces blurred and duplicated structures due to accumulated trajectory drift, while DiffRadar reconstructs a cleaner corridor layout with consistent boundaries. 
In the square-loop trajectory, baseline pipelines exhibit noticeable geometric distortions along the loop, indicating inconsistent pose estimates. 
DiffRadar instead preserves the overall loop structure and produces a more coherent radar map, demonstrating that joint pose-map optimization produces more globally consistent radar maps. 

\subsection{Controlled Stress Tests on RDST}\label{sec:results_RDST}
Beyond standard benchmarking, we evaluate DiffRadar under controlled stress conditions using the RDST dataset introduced in Section~\ref{sec:eval_setup}. 
RDST serves as a diagnostic evaluation suite that isolates radar-specific challenges, including motion degeneracy, rapid motion changes, dynamic clutter, and long-horizon trajectory drift. 
For fairness, all baselines are evaluated using their publicly released implementations with identical preprocessing pipelines, and learning-based baselines are retrained on the Radarize dataset using the same training splits. 
Across these regimes, DiffRadar consistently achieves lower trajectory error and more stable behavior than heatmap-based radar SLAM pipelines.
The results show that explicitly modeling radar sensing physics enables robust pose estimation even under motion degeneracy, dynamic interference, and long-horizon drift.

\subsubsection{Stress-Test Experiments}
We evaluate DiffRadar under progressively challenging regimes, from motion observability to dynamic interference and long-horizon consistency.

\begin{table}[t]
\centering
\caption{E1. Motion observability under zig-zag excitation. Cart-mounted and handheld trajectories.}
\label{tab:e1_corridor}
\small
\setlength{\tabcolsep}{6pt}
\renewcommand{\arraystretch}{1.12}
\begin{tabular}{@{}llcccc@{}}
\toprule
&{Method}                 & ATE$_t$\down  & ATE$_\psi$\down  & RE$_t$\down  & RE$_\psi$\down  \\
& & (m) & (rad) & (m) & (rad)\\
\midrule
\multirow{2}{*}{Cart} & Radarize           & 0.830                              & 0.161                                & \textbf{0.008}                             & 0.036                                 \\
                      & DiffRadar & \textbf{0.039}                          & \textbf{0.030}                                & 0.011                          & \textbf{0.012}                               \\ \midrule
\multirow{2}{*}{Walk} & Radarize           & 0.536                              & 0.133                                   & \textbf{0.013}                             & 0.040                                  \\
                      & DiffRadar & \textbf{0.085}                 & \textbf{0.060}                                & 0.025                          & \textbf{0.032}                               \\ \bottomrule
\end{tabular}
\end{table}

\begin{figure}
    \centering
    \includegraphics[width=0.85\linewidth]{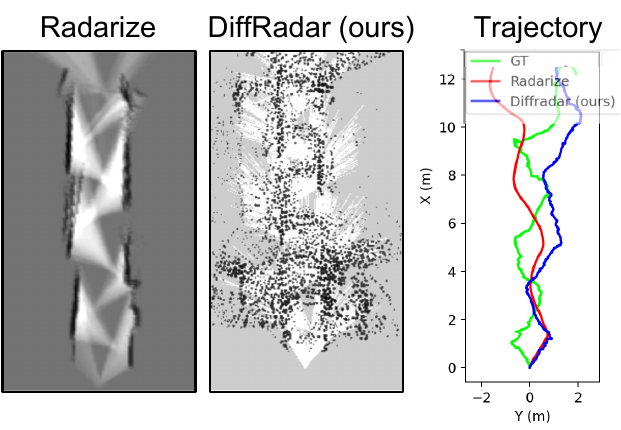}
\caption{E1. Zig-zag excitation in a corridor.
Left: Radarize reconstruction showing streak artifacts from multipath and scan degeneracy.
Middle: DiffRadar reconstruction with spatially consistent scatterers.
Right: trajectory comparison against ground truth. DiffRadar closely follows the reference trajectory while Radarize exhibits noticeable drift.}
\label{fig:e1_zigzag_results}
\Description{Two line plots for a handheld speed-transition experiment. The top plot compares ground-truth and estimated velocity profiles over time, showing the imposed motion-regime changes. The bottom plot shows translation error over time, where DiffRadar remains near zero with small fluctuations while Radarize exhibits large transient error spikes after speed changes.}
\end{figure}

\textbf{E1. Motion observability (zig-zag excitation).}
We evaluate motion observability in feature-poor corridors using scripted zig-zag trajectories that introduce lateral excitation.
Such motion exposes a common radar SLAM failure mode: when lateral structure is limited, scan-matching pipelines struggle to constrain translation.
Table~\ref{tab:e1_corridor} reports trajectory accuracy for cart-mounted and handheld runs.
DiffRadar dramatically improves global trajectory accuracy.
For cart trajectories, translational ATE decreases from 0.830\,m to 0.039\,m (over $20\times$ improvement), while for handheld runs ATE decreases from 0.536\,m to 0.085\,m.
REs remain comparable between methods.
As discussed in Section~\ref{sec:results_radarize}, Radarize's direct Doppler-based odometry can yield slightly lower short-term RE$_t$, while DiffRadar focuses on globally consistent optimization that substantially reduces trajectory drift.

Figure~\ref{fig:e1_zigzag_results} illustrates the qualitative behavior.
Radarize produces streak artifacts and noticeable trajectory drift due to scan degeneracy, whereas DiffRadar reconstructs spatially consistent scatterers and closely follows the ground-truth trajectory.
These results suggest that jointly optimizing pose and scene structure helps resolve motion ambiguity in feature-poor radar environments.

\begin{table}[t]
\centering
\caption{E2. Motion dynamics under scripted speed transitions. Cart-mounted and handheld trajectories.}
\label{tab:e2_speed}
\small
\setlength{\tabcolsep}{6pt}
\renewcommand{\arraystretch}{1.12}
\begin{tabular}{@{}llcr@{}}
\toprule
& {Method}        & \makecell{Peak post-transition \\ translation error $\downarrow$ (m)} & ATE$\downarrow$ (m) \\ \midrule
\multirow{2}{*}{Cart} & Radarize  &  1.421 &1.069                                                \\
                      & DiffRadar &   \textbf{0.206} & \textbf{ 0.148}                                               \\ \midrule
\multirow{2}{*}{Walk} & Radarize  & {7.717}                               &  {3.933}                   \\
                      & DiffRadar & \textbf{0.356} & \textbf{0.159}          \\ \bottomrule
\end{tabular}
\end{table}

\begin{figure}[t]
\begin{subfigure}{0.85\linewidth}
    \centering
    \includegraphics[width=\linewidth]{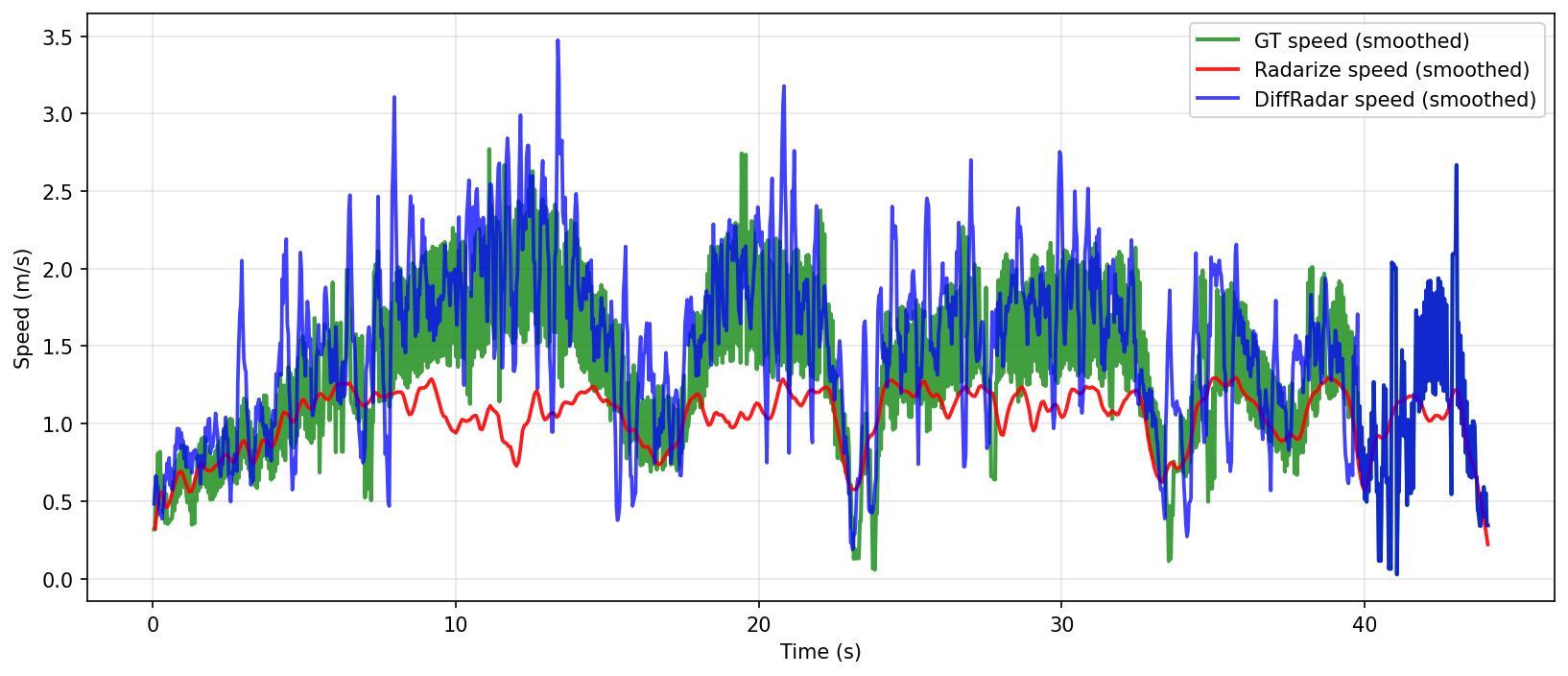}
    \caption{Ground-truth and estimated velocity profiles.}
\end{subfigure}
\vspace{0.8em}

\begin{subfigure}{0.85\linewidth}
    \centering
    \includegraphics[width=\linewidth]{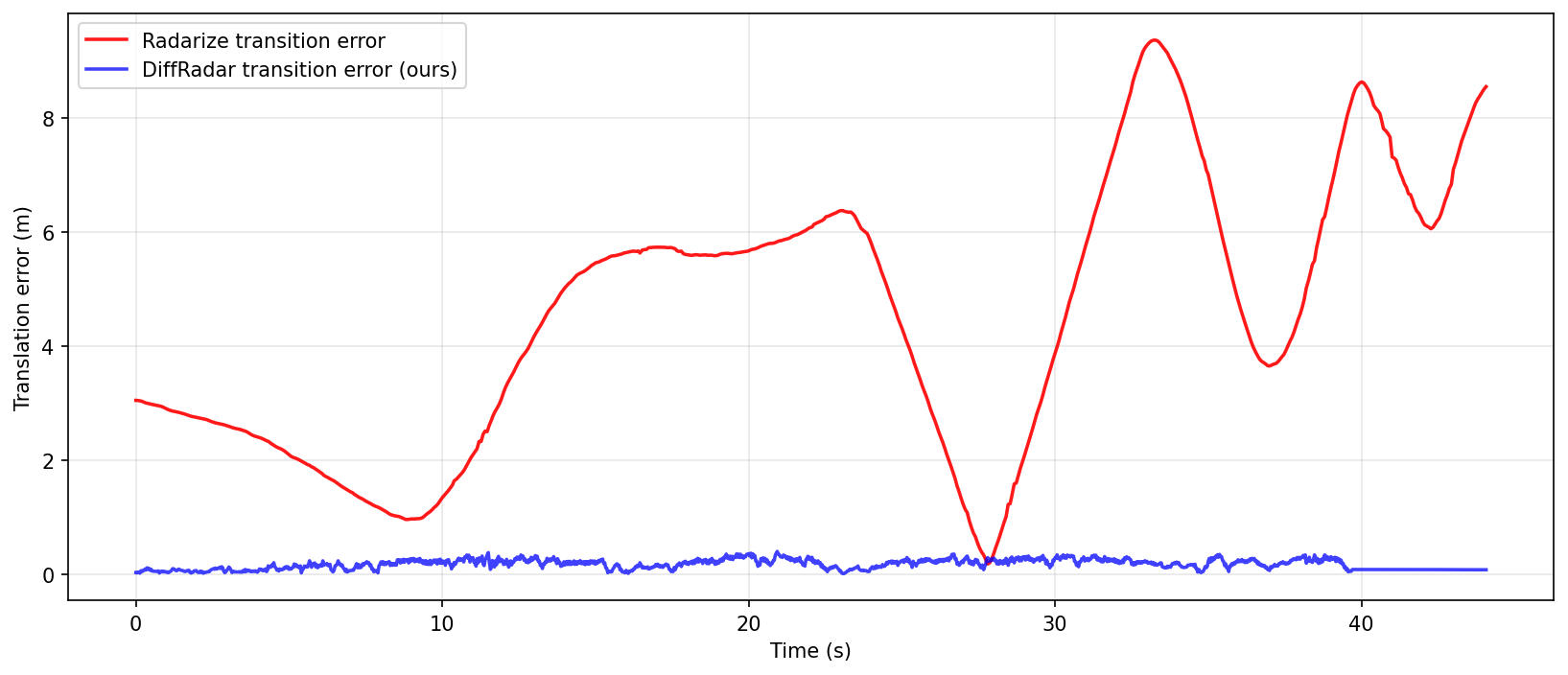}
    \caption{Translation error during motion-regime change.}
\end{subfigure}

\caption{E2. Velocity profile and translation error during handheld speed transitions.  }
\Description{Qualitative comparison for zig-zag motion in a corridor. The left panel shows a Radarize map with strong streaking and multipath artifacts, the middle panel shows a cleaner DiffRadar map with more spatially consistent scatterers, and the right panel shows trajectories against ground truth, with DiffRadar following the true path more closely and Radarize drifting away.}
\label{fig:vertical_subfigs}
\end{figure}

\textbf{E2. Motion dynamics (speed transitions).}
We evaluate robustness to motion-regime changes using scripted slow$\rightarrow$fast$\rightarrow$slow and fast$\rightarrow$slow$\rightarrow$fast schedules along identical trajectories.
Such transitions produce abrupt Doppler magnitude changes to destabilize radar odometry pipelines.
Table~\ref{tab:e2_speed} reports peak post-transition translation error and overall ATE.
DiffRadar substantially reduces transient errors following speed transitions.
For handheld trajectories, peak translation error decreases from 7.72\,m to 0.36\,m, while overall ATE drops from 3.93\,m to 0.16\,m.
Similar improvements are observed for cart-mounted runs.
These results indicate that the differentiable radar rendering model allows DiffRadar to maintain stable motion estimation under rapidly changing velocity regimes, whereas the baseline suffers large transient drift after speed transitions.
Figure~\ref{fig:vertical_subfigs} illustrates the handheld experiment.
The velocity profile confirms the imposed speed transitions, while the error plot shows that DiffRadar maintains stable estimation with significantly smaller transient spikes than Radarize.

\textbf{E3. Degenerate motion (in-place rotation).}
We evaluate the degenerate-motion regime using scripted pure rotations at fixed angular velocities.
In this setting, translational Doppler cues vanish, and orientation must be inferred purely from spatial map consistency.
Table~\ref{tab:e5_rotation} reports final yaw error and yaw drift rate.
DiffRadar significantly improves orientation estimation, reducing final yaw error from 2.69\,rad to 0.051\,rad and yaw drift from 0.098\,rad/s to 0.002\,rad/s.
This improvement arises as DiffRadar jointly optimizes pose and map structure, allowing spatial consistency constraints to stabilize orientation even when motion cues are degenerate.

\begin{table}[t]
\centering
\caption{E3. Degenerate motion under in-place rotation.}
\label{tab:e5_rotation}
\small
\setlength{\tabcolsep}{6pt}
\renewcommand{\arraystretch}{1.12}
\begin{tabular}{lrr}
\toprule
Method &
Final Yaw Error$\downarrow$ (rad) &
Yaw Drift$\downarrow$ (rad/s) \\
\midrule
Radarize    & 2.6919 & 0.098 \\
\textbf{DiffRadar} & \textbf{0.051} & \textbf{0.002} \\
\bottomrule
\end{tabular}
\end{table}

\begin{table}[t]
\centering
\caption{E4. Environmental interference under dynamic clutter (0, 2, and 4 pedestrians).}
\label{tab:e4_dynamic}
\small
\setlength{\tabcolsep}{6pt}
\renewcommand{\arraystretch}{1.12}
\begin{tabular}{@{}llrr@{}}
\toprule
 Pedestrians & Method &
ATE$_t$ $\downarrow$ (m) &
MapCons$\uparrow$ (\%) \\
\midrule
\multirow{2}{*}{0} &  Radarize    & 1.820 & 40.59 \\
& DiffRadar & \textbf{0.085} & \textbf{94.78} \\
\midrule
\multirow{2}{*}{2}  & Radarize    & 1.755 & 46.32\\
& DiffRadar & \textbf{0.104} & \textbf{93.72} \\
\midrule
\multirow{2}{*}{4}  & Radarize    & 1.344 & 47.65 \\
& DiffRadar & \textbf{0.103} & \textbf{94.90} \\
\bottomrule
\end{tabular}
\end{table}

\textbf{E4. Environmental interference (dynamic clutter).}
We evaluate robustness under dynamic interference by introducing pedestrians crossing the radar field of view along the same trajectory.
Experiments are conducted with increasing numbers of moving actors (0, 2, and 4).
Table~\ref{tab:e4_dynamic} reports trajectory error and map consistency.
DiffRadar maintains stable performance under dynamic clutter, with ATE increasing only slightly from 0.085\,m to 0.103\,m as the number of pedestrians increases.
In contrast, Radarize exhibits substantially larger trajectory errors across all settings.
DiffRadar also preserves high map consistency (approximately 94\%), whereas Radarize produces maps with significantly lower structural consistency (40–47\%).
These results suggest that the visibility-conditioned radar field representation helps suppress transient scatterers caused by moving objects.

\begin{table}[t]
\centering
\caption{E5. Long-horizon consistency under loop closures (>100\,m). Cart-mounted \& handheld trajectories.}
\label{tab:e3_loops}
\small
\setlength{\tabcolsep}{6pt}
\renewcommand{\arraystretch}{1.12}
\begin{tabular}{@{}llrr@{}}
\toprule
& {Method} &
Loop Drift$\downarrow$ (m/100m) &
ATE$\downarrow$ (m) \\
\midrule
\multirow{2}{*}{Cart} & Radarize    & 20.438 & 2.921 \\
                     & DiffRadar & \textbf{0.511} & \textbf{0.017} \\
\midrule

\multirow{2}{*}{Walk} & Radarize    & 26.438 & 6.295 \\
                    & DiffRadar & \textbf{1.385} & \textbf{0.031} \\
\bottomrule
\end{tabular}
\end{table}

\textbf{E5. Long-horizon consistency (loop closures).}
We evaluate long-horizon consistency using trajectories with loops exceeding 100\,m and repeated revisits.
Table~\ref{tab:e3_loops} reports loop drift and trajectory ATE for cart-mounted and handheld runs.
DiffRadar dramatically reduces global drift.
For cart-mounted runs, loop drift decreases from 20.44\,m/100\,m to 0.51\,m/100\,m, while ATE drops from 2.92\,m to 0.017\,m.
Similar improvements are observed for handheld trajectories. 
These results indicate that jointly optimizing pose and scene structure enables DiffRadar to maintain globally consistent trajectories over long horizons, whereas scan-matching pipelines accumulate substantial drift.

\begin{table}[!tbp]
\centering
\caption{End-to-end SLAM trajectory accuracy on RDST.}
\label{tab:slam_traj_RDST}
\small
\setlength{\tabcolsep}{6pt}
\renewcommand{\arraystretch}{1.15}
\begin{tabular}{lrrrr}
\toprule
Method & ATE$_t$\down (m) & ATE$_\psi$\down (rad) & RE$_t$\down (m) & RE$_\psi$\down (rad) \\
\midrule
Radarize    & 0.823 & 0.162 & \textbf{0.024} & 0.172 \\
\midrule
{DiffRadar} & \textbf{0.129} & \textbf{0.051} & 0.036& \textbf{0.035} \\
\bottomrule
\end{tabular}
\end{table}

\begin{table}[!tbp]
\centering
\caption{Map fidelity and artifact metrics on RDST.}
\label{tab:map_quality_RDST}
\small
\setlength{\tabcolsep}{6pt}
\renewcommand{\arraystretch}{1.15}
\begin{tabular}{lrrr}
\toprule
Method & Loop Drift\down & MapCons\up (\%) & V-Artifact\down (\%) \\
& (m/100m) &  &  \\
\midrule
Radarize & 0.35 & 42.59 & 53.74 \\
{DiffRadar} & \textbf{0.18}& \textbf{94.78} & \textbf{38.61}  \\
\bottomrule
\end{tabular}
\end{table}
\subsubsection{Overall SLAM Performance on RDST}
Beyond regime-specific stress tests, we evaluate overall SLAM performance across the RDST dataset.
Table~\ref{tab:slam_traj_RDST} summarizes trajectory accuracy averaged across all sequences.
DiffRadar substantially improves global localization accuracy, achieving over a $6\times$ reduction in translational ATE and a $3\times$ reduction in rotational ATE compared with Radarize.
Although Radarize achieves slightly lower relative translation error (RE$_t$) due to its direct Doppler-based odometry estimation, DiffRadar significantly reduces accumulated trajectory drift through joint pose--map optimization.
We further evaluate map fidelity and structural consistency.
As shown in Table~\ref{tab:map_quality_RDST}, DiffRadar more than doubles map consistency (43\% $\rightarrow$ 95\%) while also reducing loop drift and vertical artifacts.
These results indicate that the robustness gains observed in the controlled stress tests translate to more accurate trajectories and structurally consistent maps in realistic deployment scenarios.

\subsection{Runtime and Ablation Analysis}
\label{sec:results_ablation_perf}

\textbf{Runtime Analysis.}
We evaluate the runtime efficiency of DiffRadar and compare it with Radarize.
Table~\ref{tab:runtime} reports steady-state system performance measured on the Radarize dataset after map stabilization.
FPS reflects end-to-end SLAM throughput after initialization warm-up. 
DiffRadar achieves real-time performance while maintaining higher accuracy.
In steady state, DiffRadar processes frames at 70 FPS compared with 55 FPS for Radarize.
Despite optimizing a differentiable radar field representation, the system remains efficient due to two design choices: (i) the compact Gaussian scene representation, which avoids storing dense radar heatmaps, and (ii) a fixed-lag optimization window that bounds per-frame computation.
DiffRadar also maintains a substantially smaller memory footprint, requiring only 40\,MB for the map compared with 162\,MB for Radarize.

Importantly, this efficiency remains stable over long trajectories.
As demonstrated in the RDST long-horizon experiments (Table~\ref{tab:e3_loops}), DiffRadar maintains globally consistent trajectories on loops exceeding 100\,m while operating in real time.
These results indicate that DiffRadar scales well with trajectory length and remains practical for extended deployments.

\begin{table}[t]
\centering
\caption{Steady-state runtime comparison. Initialization denotes offline preprocessing before SLAM execution. FPS measures end-to-end SLAM throughput after map stabilization.}
\label{tab:runtime}
\small
\setlength{\tabcolsep}{4pt}
\renewcommand{\arraystretch}{1.1}
\begin{tabular}{lcrr}
\toprule
Method & Initialization Time (hrs) & FPS$\uparrow$ & Map Size (MB) \\
\midrule
Radarize & 3.56 & 55 & 162 \\
DiffRadar          & \textbf{1.35} & \textbf{70} & \textbf{40} \\
\bottomrule
\end{tabular}
\end{table}

\begin{table}[t]
\centering
\caption{{Ablation study on the public benchmark.}}
\label{tab:ablations}
\small
\setlength{\tabcolsep}{6pt}
\renewcommand{\arraystretch}{1.12}
\begin{tabular}{lrrrr}
\toprule
Method & ATE$_t$\down & ATE$_\psi$\down & RE$_t$\down & RE$_\psi$\down  \\
& (m) & (rad)& (m) &(rad) \\

\midrule
Full DiffRadar & \textbf{0.034}  & \textbf{0.054 }& 0.020 &\textbf{ 0.022} \\
No Doppler residual &  0.078 & 0.264 & 0.011 & 0.033 \\
No RA residual &  0.093 & 0.067 & \textbf{0.007} & 0.066 \\
No Anisotropic covariance & 0.085  & 0.135 & 0.013 & 0.032 \\
No Visibility gating & 0.085  & 0.134 & 0.012 & 0.027 \\
Calibration $\pm$5\% & 0.092 & 0.156 & 0.013 & 0.029 \\
\bottomrule
\end{tabular}
\end{table}

\textbf{Ablation Study.}
We analyze how key design components affect SLAM accuracy using the Radarize benchmark.
Table~\ref{tab:ablations} reports trajectory accuracy after removing individual components from the full system.
The ablation highlights the complementary roles of Doppler and spatial radar measurements.
Removing Doppler residuals substantially degrades translational accuracy, while removing RA residuals degrades both translation and orientation estimation, confirming that Doppler cues and spatial radar structure jointly constrain motion.
Physics-aware modeling further stabilizes optimization: replacing anisotropic covariance with isotropic covariance or removing visibility gating both increase trajectory errors, indicating that modeling radar sensing anisotropy and enforcing physically consistent visibility are important for accurate pose–map optimization.
DiffRadar also remains robust to moderate calibration noise, as perturbing radar calibration parameters ($D,B$) by $\pm5\%$ causes only moderate degradation.

Some variants exhibit slightly lower RE despite worse global accuracy.
This occurs because removing constraints can locally smooth short-term motion estimates while introducing larger long-horizon drift, which is reflected in increased ATE.
Overall, the results confirm that each component contributes to accurate and stable radar SLAM.

\section{Related Work} \label{sec:related}

Prior work on radar-based mapping can be broadly divided into three categories:
(i) radar SLAM and odometry systems,
(ii) learning-based radar perception, and
(iii) differentiable scene representations for radar sensing.

\textbf{Radar SLAM and Doppler Odometry.}
Radar SLAM systems typically estimate motion by matching scans or features extracted from range--azimuth measurements.
Early work demonstrated radar-only ego-motion estimation using scan registration or graph matching between radar returns \cite{cen2018precise,cen2019radar,checchin2010Radar,hong2021radar}.
Subsequent systems improved robustness through probabilistic estimation, learned feature extraction, and outlier-resistant odometry pipelines \cite{barnes2020Masking,burnett2021radar,lim2023orora,adolfsson2021CFEAR,hilger2024randt}.
More recent approaches incorporate Doppler measurements for improved motion estimation \cite{sie2024radarize,han2025equiro,burnett2021Do,legentil2025DRO,kubelka2024Do}, while radar--inertial systems fuse radar sensing with IMU measurements for improved trajectory estimation \cite{amodeo2024gaussianrio}.
However, most prior pipelines operate on discretized radar images or point representations and rely on scan matching rather than a differentiable forward model \cite{sie2024radarize,lu2020milliego,prabhakara2023high,legentil2025DRO}.

\textbf{Learning-Based Radar Perception.}
Deep learning has been widely applied to radar perception and odometry.
Many approaches estimate ego-motion from range--Doppler or radar tensor representations using convolutional or recurrent neural networks \cite{lu2020milliego,lu2020see,sie2024radarize}.
Other works attempt to improve spatial resolution or densify radar observations through learned representations and generative models, such as diffusion-based methods that jointly infer scene structure and motion from radar measurements \cite{wang2025sddiff,prabhakara2023high}.
These approaches typically rely on supervised training and treat the radar sensing process implicitly.

\textbf{Differentiable Scene Representations.}
Differentiable scene representations such as neural radiance fields and 3D Gaussian Splatting enable continuous modeling of scene geometry and sensor poses through differentiable rendering \cite{kerbl20233d,keetha2024splatam,matsuki2024gaussian,zhang2026rf}.
Recent work has extended neural or Gaussian representations to radar sensing, including neural radar fields and SAR imaging models \cite{rafidashti2025neuradar,lei2024sar,zhang2025rf4d,giacomini2025splat}, as well as Gaussian-based radar perception and sensor fusion methods \cite{kung2025radarsplat,xiao2025radgs,montiel2026gaussiancar,xiao2024liv}.
However, these systems focus primarily on rendering, reconstruction, or multi-sensor perception rather than radar-only SLAM grounded in a physics-consistent radar forward model.
\textbf{In contrast}, DiffRadar integrates an FMCW-consistent radar forward model into a differentiable Gaussian scene representation, enabling joint optimization of robot pose and scene structure directly from radar measurements \cite{keetha2024splatam,matsuki2024gaussian}.
\section{Conclusion and Discussion}

\textbf{Conclusion.}
We presented DiffRadar, a real-time radar SLAM system that models radar sensing as a differentiable physics-aware Gaussian field. 
By jointly optimizing sensing physics, scene structure, and ego-motion within a unified representation, DiffRadar enables robust radar-native SLAM while maintaining real-time performance on commodity mmWave hardware. 
Experiments across benchmarks and controlled stress tests demonstrate substantial improvements in accuracy and robustness under challenging conditions. 
These results suggest that integrating radar sensing physics with differentiable scene representations provides a promising direction for next-generation radar SLAM systems.

\textbf{Limitations.}
Despite these improvements, several deployment and evaluation considerations remain: 
(i) Our evaluation uses a specific commodity FMCW radar configuration with a fixed operating range, antenna layout, and sensing parameters; adapting DiffRadar to substantially different radar hardware or sensing settings may require additional calibration or parameter tuning. 
(ii) Although DiffRadar shows robustness under moderate dynamic clutter (Section~\ref{sec:results_RDST}), environments with dense moving objects may still produce inconsistent radar returns that affect pose estimation.



\bibliographystyle{acm}

\bibliography{bib/refs2}

\onecolumn


\end{document}